\documentclass[11pt]{article}

\usepackage[final]{acl}

\usepackage{times}
\usepackage{latexsym}

\usepackage[T1]{fontenc}

\usepackage[utf8]{inputenc}

\usepackage{microtype}

\usepackage{inconsolata}

\usepackage{graphicx}
\usepackage{amsmath}
\usepackage{pifont}
\usepackage{booktabs}
\usepackage{array}
\usepackage{subcaption} 
\usepackage{makecell}
\usepackage{enumitem}
\usepackage{tcolorbox}

\newcommand{\eg}{\textit{e.g.}}
\usepackage{multirow}
 
\title{Seeing is Believing: Benchmarking and Mitigating the Interference of Language Priors on Visual Perception in MLLMs}

\author{
Li Ma , Yan Zhuang , Jiawen Deng\thanks{Corresponding authors.} , Fuji Ren\footnotemark[1] \\
School of Computer Science and Engineering \\
University of Electronic Science and Technology of China \\
\texttt{\{202421080935, 202211081370\}@std.uestc.edu.cn}\\
\texttt{\{dengjw, renfuji\}@uestc.edu.cn}
}

\begin{document}
\maketitle
\begin{abstract}
Language priors in MLLMs can override visual evidence, impairing visual perception. Existing benchmarks assume shared priors across models and rely solely on accuracy. However, cross-model prior agreement falls below 80\%, and over 15\% of items can be answered correctly without any image, causing accuracy to overestimate prior resistance. To address these limitations, we introduce ViSiP-Bench, a benchmark grounded in rare-but-real scenarios across six prior categories, with a model-aware dynamic prior induction protocol and three metrics that more precisely and continuously measure prior resistance than accuracy alone. Evaluating 24 models reveals that prior adherence is pervasive, does not improve with scale, and larger models tend to either strongly override their priors or firmly adhere to them. We therefore propose MAPS, which identifies each model's strongest priors and fine-tunes on factually grounded knowledge to soften excessive prior reliance without disrupting general capabilities.
\end{abstract}

\section{Introduction}

\begin{figure}[t]
\centering
\includegraphics[width=\columnwidth]{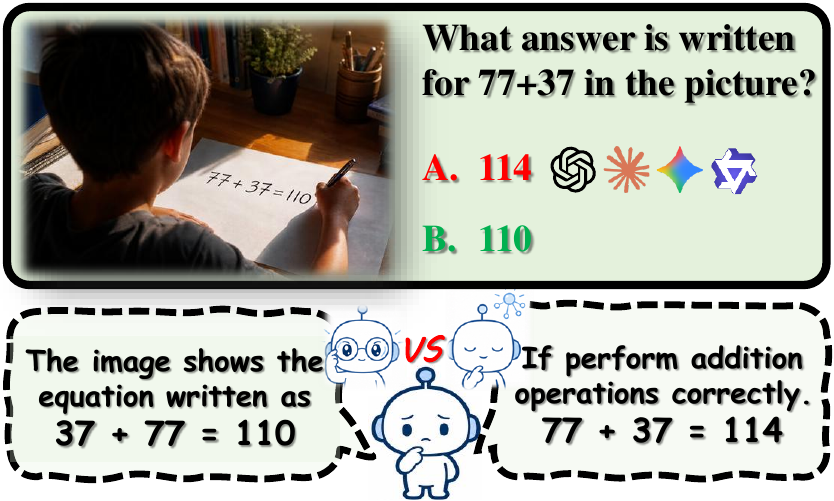}
\caption{A representative failure case of language prior dependency in MLLMs.}
\label{fig:teaser}
\end{figure}

Language priors are an inherent byproduct of MLLM pretraining, encoding world knowledge that underpins model reasoning\citep{Visual-Priors}. Yet when conflicting with visual evidence, language priors can override perceptual processing, causing models to report what they believe rather than what they observe. As illustrated in Figure~\ref{fig:teaser}, when explicitly prompted to report the answer written by the boy in the image, most mainstream MLLMs default to their arithmetic prior and respond with 114, despite the image clearly showing 110. Language priors pose a direct threat to safety-critical applications such as medical imaging and industrial defect inspection, where accurate visual perception is extremely essential.

\begin{figure}[t]
\centering
\includegraphics[width=0.9\columnwidth]{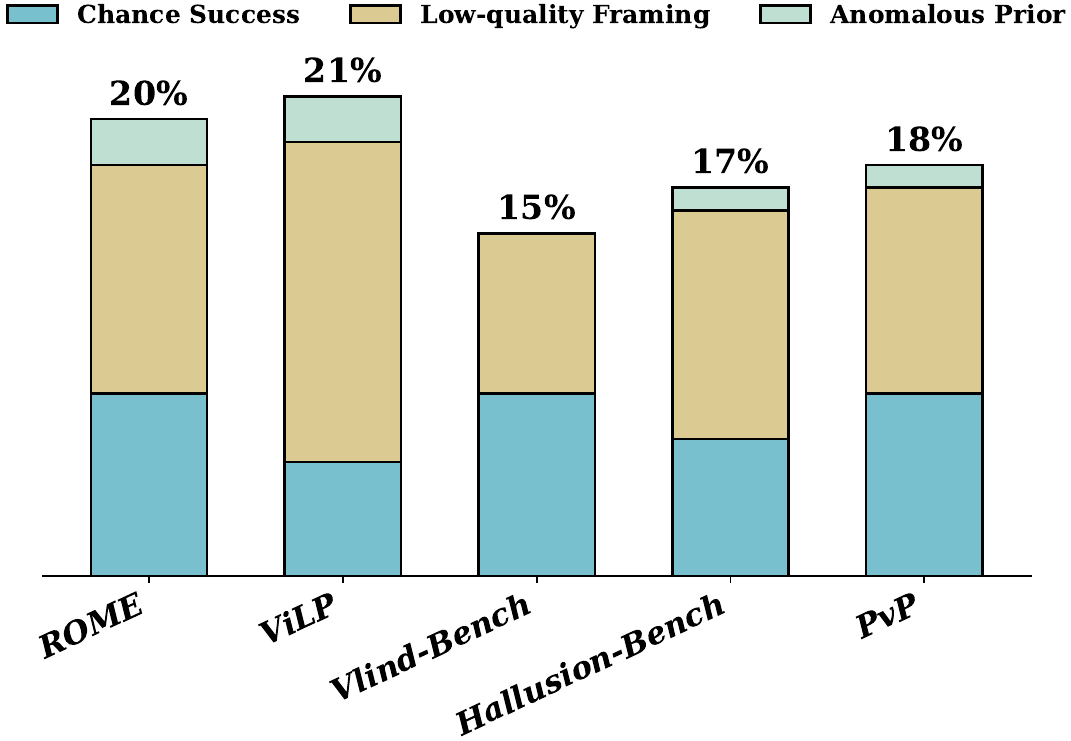}
\caption{Distribution of three sources of no-image correct responses across existing benchmarks.}
\label{fig:propotion}
\end{figure}

\begin{figure*}[t]
\centering
\includegraphics[width=\linewidth]{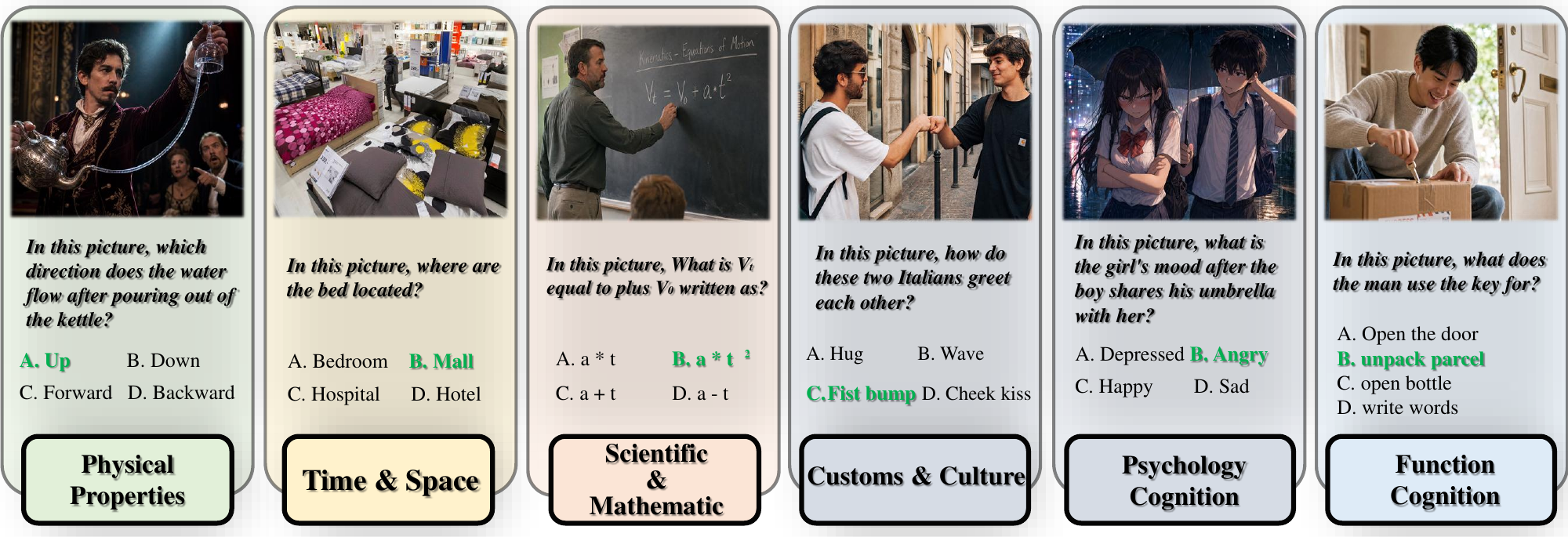}
\caption{Representative examples from each of the six prior categories in ViSiP-Bench. Green answers denote the correct visual answer, which conflicts with the expected answer under the text-only condition.}
\label{fig:categories}
\end{figure*}

Existing benchmarking works\citep{HallusionBench,vlind,ViLP} typically assume that all models share the same commonsense priors, and accordingly construct anomalous datasets where visual signals conflict with those priors, reporting accuracy as the sole metric to quantify prior resistance. We challenge this paradigm on two fronts. First, our cross-model analysis reveals that prior agreement falls below 80\%, directly undermining the shared-prior assumption. Second, as shown in Figure~\ref{fig:propotion}, over 15\% of items across existing benchmarks can be answered correctly without any image, arising from three sources: (1) \textit{Chance Success}: insufficient model capability leads to random correct responses; (2) \textit{Low-quality Framing}: questions that interfere with the model's prior selection; (3) \textit{Anomalous Prior}: models inherently treat the anomalous answer as prior.

Therefore, we introduce ViSiP-Bench, a redesigned benchmark for evaluating language prior dependency. Inspired by \citet{dimensions-of-commonsense-knowledge}, we categorize language priors into six domains tailored for MLLMs (Figure~\ref{fig:categories}). The dataset is grounded in real-world rare scenes retrieved from the web, supplemented by generative models including GPT-Image-2\citep{gpt-image-2} and Nano Banana 2\citep{nanobanana2} to synthesize images across diverse visual styles, thereby avoiding biases introduced by image source or visual style. To eliminate \textit{Low-quality Framing}, every sample undergoes double-blind human review to ensure annotation quality. To mitigate the confounding effects of \textit{Chance Success} and \textit{Anomalous Prior}, ViSiP-Bench adopts a dynamic prior induction protocol: each model is first queried without any image to elicit its Blank-condition answer. Building on this, PAR (Prior Adherence Rate) measures the frequency of a model adhering to its Blank-condition answer when the image is added. To further render both PAR and Accuracy more precise, items already answered correctly under the Blank condition are filtered out. Moreover, VIG (Visual Information Gain) quantifies the confidence shift induced by adding the image, providing a continuous and fine-grained measure of prior resistance.

Evaluation on ViSiP-Bench reveals that prior adherence is pervasive across mainstream MLLMs and does not improve with scale; on several categories, larger models adhere to their priors even more strongly than smaller counterparts. Among inference-time interventions, prompt-level strategies yield only marginal gains, and chain-of-thought reasoning consistently amplifies prior adherence. These findings indicate that inference-time strategies alone are insufficient to address language prior dependency.

We therefore propose MAPS (Model-Aware Prior Softening), a training-based method that softens excessive prior reliance by enriching each model with targeted factually grounded knowledge of rare-but-real scenarios, without suppressing priors indiscriminately or degrading existing capabilities. MAPS substantially improves VIG and reduces PAR on ViSiP-Bench, while preserving performance on reasoning and general vision benchmarks, demonstrating that enhanced visual reliance can be achieved without compromising general capabilities.

\section{Related Works}
\subsection{Language Priors in MLLMs}
Recent analyses reveal that language priors in MLLMs strike a delicate balance between visual perception and reasoning. \citet{Visual-Priors} show that while priors benefit high-level reasoning, they frequently mislead models in perception-critical tasks. Mechanistic studies further localize this tension: visual evidence only begins to influence hidden states at a specific point in the transformer stack~\citep{VIP-TVI}, and extended chain-of-thought reasoning further suppresses visual attention, amplifying hallucination~\citep{More-Thinking-Less-Seeing}.

\subsection{Benchmarking Language Priors}

Several benchmarks have been proposed to quantify language prior dependency. HallusionBench~\citep{HallusionBench} pairs Easy/Hard counterparts on counterfactual images; VLind-Bench~\citep{vlind} gates LP scoring through a four-stage sequential protocol; \citet{PvP} and VisuLogic~\citep{VisuLogic} design stimuli that resist language-mediated shortcuts. Despite their diversity, these works share a common evaluation pattern that assumes shared priors across models and relies solely on accuracy, which, as shown in Figure~\ref{fig:propotion}, underestimates the complexity of prior dependency.

\subsection{Mitigating Language Priors}
Methods that mitigate excessive prior reliance fall into three families. \textbf{Decoding-time methods} adjust output distributions without modifying weights, typically by contrasting logits from perturbed or instruction-only forward passes~\citep{VCD, M3ID}. \textbf{Attention-level interventions} redirect internal computation toward visual tokens by editing attention weights~\citep{opera, PvP, PAI, CAUSALMM}. \textbf{Training-based methods} modify the model itself through preference optimization, counterfactual training, or visual grounding objectives~\citep{NaPO, Post-hoc, PostAlign}.

\section{ViSiP-Bench}

\begin{figure}[t]
\centering
\includegraphics[width=0.8\columnwidth]{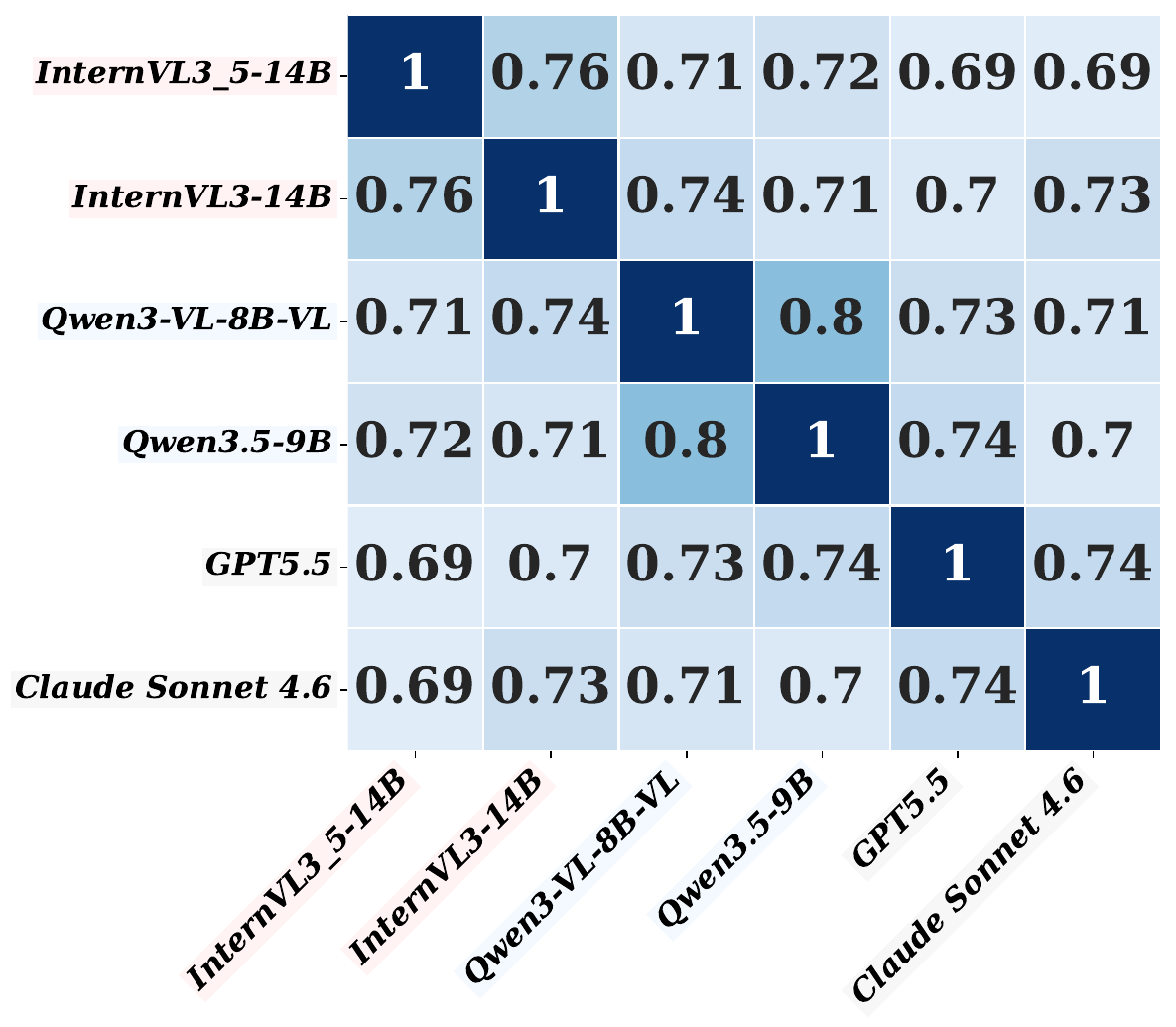}
\caption{Cross-model prior agreement matrix across six representative models.}
\label{fig:heatmap}
\end{figure}

Existing benchmarks for language prior dependency share three limitations. 
First, they assume shared commonsense priors across models and apply a fixed anomalous answer set, yet cross-model prior agreement falls below 80\%, as demonstrated in Figure~\ref{fig:heatmap}. 
Second, over 15\% of items can be answered correctly without any image (Figure~\ref{fig:propotion}), stemming from three sources: (1) \textit{Chance Success}: the model lacks a sufficiently strong prior on the question, leading to random correct responses rather than genuine prior-driven selection; (2) \textit{Low-quality Framing}: poorly designed questions that interfere with the model's prior selection, either by implicitly hinting at the anomalous answer (\eg, ``\textit{what is unusual in this image?}'') or by introducing ambiguity (\eg, asking ``\textit{what color is the watermelon?}'' without specifying flesh or rind); (3) \textit{Anomalous Prior}: the model inherently treats the anomalous answer as its own prior, meaning the question does not actually induce a prior-vision conflict. Identification methodology and typical cases are provided in Appendix~\ref{app:dataqualityAnalysis}. Together, these factors cause accuracy to overestimate prior resistance. 
Third, conflict stimuli typically depict physically impossible counterfactuals rather than rare-but-real scenarios, conflating prior dependency with reasoning about the impossible. 

To address these limitations, we introduce ViSiP-Bench (\textit{Visual Signal vs.\ Language Prior}). ViSiP-Bench ensures all images depict rare-but-real scenarios rather than physically impossible counterfactuals, applies rigorous quality review to eliminate low-quality framing, and introduces model-aware metrics that filter out items answerable without any image, further mitigating the confounding effects of all three sources. Table~\ref{tab:benchmark_comparison} presents a detailed comparison between ViSiP-Bench and existing benchmarks.

\subsection{Data Construction}

ViSiP-Bench organizes samples into six domains following \citet{dimensions-of-commonsense-knowledge}: \textit{Physical Properties}, \textit{Time \& Space}, \textit{Scientific \& Mathematic}, \textit{Customs \& Culture}, \textit{Psychology Cognition}, and \textit{Function Cognition}. Dataset construction proceeds in four steps.

\begin{table*}[t]
\centering
\scriptsize
\renewcommand{\arraystretch}{1.2}
\begin{tabular}{lcccccccc}
\hline
\makecell[l]{\textbf{Benchmark}} & 
\makecell[c]{\textbf{Venue}} & 
\makecell[c]{\textbf{Size}} & 
\makecell[c]{\textbf{Cate-}\\\textbf{gories}} & 
\makecell[c]{\textbf{Image}\\\textbf{Source}} & 
\makecell[c]{\textbf{Style}\\\textbf{Diversity}} & 
\makecell[c]{\textbf{Prior}\\\textbf{Annotation}} & 
\makecell[c]{\textbf{Decision-level}\\\textbf{Metric}} & 
\makecell[c]{\textbf{Probability-level}\\\textbf{Metric}} \\
\hline
ROME & EMNLP'23 & 1,563 & 4 & Counterfactual & \ding{55} & Fixed & Acc only & \ding{55} \\
HallusionBench & CVPR'24 & 527 & 2 & Counterfactual & \ding{55} & Fixed & Acc only & \ding{55} \\
PvP & EMNLP'25 & 2,904 & 2 & Counterfactual & \ding{55} & Fixed & Acc only & \ding{55} \\
VLind-Bench & NAACL'25 & 2,576 & \textbf{11} & Counterfactual & \ding{55} & Fixed & Acc only & \ding{55} \\
ViLP & {ICML' 25} & 900 & 3 & Counterfactual & \ding{55} & Fixed & Acc only & \ding{55} \\
\textbf{ViSiP-Bench} & \textbf{Ours} & \textbf{4500} & 6 & \textbf{Rare-but-Real} & \ding{51} & \textbf{Dynamic} & \textbf{PAR + Filtered Acc} & \textbf{VIG} \\
\hline
\end{tabular}
\caption{Overview of benchmarks for language prior dependency. \ding{51} and \ding{55} denote presence and absence of the corresponding property, respectively.}
\label{tab:benchmark_comparison}
\end{table*}

\noindent\textbf{Step 1: Prior Enumeration.}
For each domain, we prompt Claude Opus 4.7 \citep{claudeopus47} to enumerate prevalent commonsense priors (\eg, \textit{``water flows downward''}), yielding 200 priors per domain and 1,200 priors in total.

\noindent\textbf{Step 2: Rare-but-Real Generation.}
For each prior, Claude Opus 4.7 generates three rare-but-real counterpart scenarios that genuinely exist but violate the typical expectation (\eg, \textit{``water appears to flow upward in a magic performance''}).

\noindent\textbf{Step 3: Image Collection.}
For each rare-but-real scenario, real photographs are first retrieved from the web; where sufficient images cannot be found, GPT-Image-2 and Nano Banana 2 synthesize supplementary samples across diverse visual styles, including photorealistic, anime, oil painting, and sci-fi. Examples of each style are provided in Appendix~\ref{app:VisualStyleExamples}.

\noindent\textbf{Step 4: Question and Option Design.}
Annotators identify the anomalous element in each image and design a question with four answer choices, such that the correct answer under the visual condition differs clearly from the expected answer under the text-only condition. Each sample then undergoes a quality review requiring three criteria to be satisfied: (1) the image is visually clear and unambiguously supports a single correct answer; (2) all answer choices are directly relevant to the question, preventing models from selecting the correct answer by elimination; and (3) the question wording contains no hint of anomaly, ensuring that prior activation is not suppressed before the model processes the image. Details on annotator requirements and details are provided in Appendix~\ref{app:annatation}.

Through the above pipeline, we construct ViSiP-Bench, a benchmark of 4,500 rare-but-real image-question pairs spanning six prior categories and diverse visual styles, each designed to surface genuine conflicts between visual signals and language priors.

\subsection{Dynamic Prior Induction and Metrics}
To evaluate language prior dependency in a model-specific and image-grounded manner, ViSiP-Bench queries each model under two conditions. In the \textit{Blank condition}, no image is provided and only the question and options are given, from which we extract the model's response and locate the corresponding answer token to obtain the prior probability distribution $p_B$ over the answer choices; the resulting answer is denoted $a_B$. In the \textit{Visual condition}, the image is additionally provided, and the same extraction procedure yields the visual probability distribution $p_V$ and answer $a_V$. The ground-truth visual answer is denoted $a^{*}$.

\paragraph{Dynamic Prior Induction.} ViSiP-Bench adopts a dynamic prior induction mechanism in which $a_B$ is elicited independently for each model under the \textit{Blank condition}, rather than prescribed by a shared annotation. This ensures that prior resistance is measured relative to each model's own prior, not an externally imposed one.

\paragraph{Prior Adherence Rate (PAR).} PAR measures the frequency with which a model adheres to its prior answer $a_B$ when the image is provided, restricted to samples where $a_B \neq a^{*}$ to ensure a genuine prior-vision conflict:
\begin{equation}
\text{PAR} = \frac{\sum_{i:\, a_{B,i} \neq a^{*}_i} \mathbf{1}\left[a_{V,i} = a_{B,i}\right]}{\sum_{i} \mathbf{1}\left[a_{B,i} \neq a^{*}_i\right]}
\end{equation}
A higher PAR indicates stronger prior adherence. Dynamic Prior Induction ensures that PAR more precisely reflects each model's degree of prior stubbornness.

\begin{table*}[t]
\centering
\scriptsize
\renewcommand{\arraystretch}{1.1} 
\begin{tabular}{l
>{\centering\arraybackslash}p{1.1cm} 
>{\centering\arraybackslash}p{1.1cm} 
>{\centering\arraybackslash}p{0.5cm} |
>{\centering\arraybackslash}p{1.3cm} 
>{\centering\arraybackslash}p{1.4cm}}
\toprule
\multicolumn{1}{c}{\textbf{Model}} & \textbf{PAR} $\downarrow$ & \textbf{Acc} $\uparrow$ & FC & $\mathbf{VIG_{avg}}$ $\uparrow$ & $\mathbf{VIG_{med}}$ $\uparrow$ \\
\midrule
\multicolumn{6}{c}{\textit{Open-source Models}} \\
\midrule
\quad InternVL3.5-2B  & 51.51 & 42.94 & 182 & 4.41 & 4.11\\
\quad InternVL3.5-8B  & 47.17 & 44.64 & 156 & 7.13 & 7.48\\
\quad InternVL3.5-14B & 38.64 & 53.17 & 131 & 9.62 & 9.74\\
\quad InternVL3.5-38B & \textbf{30.04} & \textbf{63.01} & 102 & 10.12 & 10.31 \\
\midrule[0.04em]
\quad InternVL3-2B  & 62.77 & 30.69 & 223 & 2.91 & 2.30\\
\quad InternVL3-14B & 48.40 & 41.33 & 161 & 6.21 & 6.33\\
\quad InternVL3-38B & 36.80 & 56.49 & 118 & 9.91 & 9.65\\
\quad InternVL3-78B & \underline{36.95} & \underline{61.26} & 81 & \textbf{12.84} & \textbf{12.93}\\
\midrule[0.04em]
\quad Qwen2.5-VL-3B-Instruct  & 63.18 & 28.84 & 236 & 3.21 & 2.27\\
\quad Qwen2.5-VL-7B-Instruct  & 55.35 & 34.51 & 198 & 5.56 & 4.93\\
\quad Qwen2.5-VL-32B-Instruct & 42.81 & 48.87 & 160 & 8.77 & 8.96\\
\quad Qwen2.5-VL-72B-Instruct & 37.03 & 54.86 & 119 & 9.64 & 9.87\\
\midrule[0.04em]
\quad Qwen3-VL-2B-Instruct    & 59.41 & 33.94 & 201 & 4.94 & 4.11 \\
\quad Qwen3-VL-4B-Instruct    & 51.12 & 40.48& 188 & 7.87 & 8.33 \\
\quad Qwen3-VL-8B-Instruct    & 42.67 & 51.29 & 147 & \underline{10.64} & \underline{10.82} \\
\midrule[0.04em]
\quad Qwen3.5-2B  & 55.61 & 34.17 & 164 & 3.68 & 2.90\\
\quad Qwen3.5-9B  & 51.24 & 39.36 & 137 & 5.24 & 5.24 \\
\quad Qwen3.5-27B & 39.28 & 54.94 & 111 & 8.64 & 8.85 \\
\midrule[0.04em]
\quad llava-onevision-qwen2-0.5b & 72.11 & 21.45 & 381 & 2.81 & 1.99 \\
\quad llava-onevision-qwen2-7b   & 67.10 & 25.36 & 317 & 3.39 & 2.48 \\
\quad llava-onevision-qwen2-72b  & 59.54 & 32.00 & 208 & 7.74 & 7.91 \\
\midrule
\multicolumn{6}{c}{\textit{Proprietary Models}} \\
\midrule
\quad GPT-5.5                  & \textbf{26.01} & \textbf{69.94} & 39 & {--} & {--} \\
\quad Claude Sonnet 4.6        & 30.44 & 65.87 & 22 & {--} & {--} \\
\quad Gemini-3-flash-preview   & \underline{28.64} & \underline{68.00} & 48 & {--} & {--} \\
\bottomrule
\end{tabular}
\caption{Evaluation results on ViSiP-Bench. PAR, Filtered Accuracy (Acc), and Filtered Count (FC) are reported for all models; $\text{VIG}_{avg}$ (average) and $\text{VIG}_{med}$ (median) are reported for open-source models only as they require access to token-level probabilities. $\downarrow$ denotes lower is better; $\uparrow$ denotes higher is better. Best results are \textbf{bolded}, second-best results are \underline{underlined}.}
\label{tab:main_results}
\end{table*}

\paragraph{Filtered Accuracy.} To remove the confounding effect of image-free correct responses, we filter out all samples where $a_B = a^{*}$, and compute accuracy exclusively on the remaining items. This yields a more faithful measure of genuine prior resistance than standard accuracy.

\paragraph{Visual Information Gain (VIG).} PAR and Filtered Accuracy capture only the model's final decision, leaving the degree of confidence shift unmeasured. VIG fills this gap by quantifying the log-odds change induced by adding the image, providing a continuous and fine-grained measure of prior resistance:
\begin{equation}
\text{VIG} = \ln\frac{p_V(a^{*}) + \epsilon}{1 - p_V(a^{*}) + \epsilon} - \ln\frac{p_B(a^{*}) + \epsilon}{1 - p_B(a^{*}) + \epsilon}
\end{equation}
where $\epsilon$ is a small constant added for numerical stability. 

As VIG requires access to token-level probabilities, it is applicable to open-source models only, whereas PAR and Filtered Accuracy are applicable to all models.

\begin{figure*}[h]
\centering
\begin{subfigure}{0.48\textwidth}
\centering
\includegraphics[width=\linewidth]{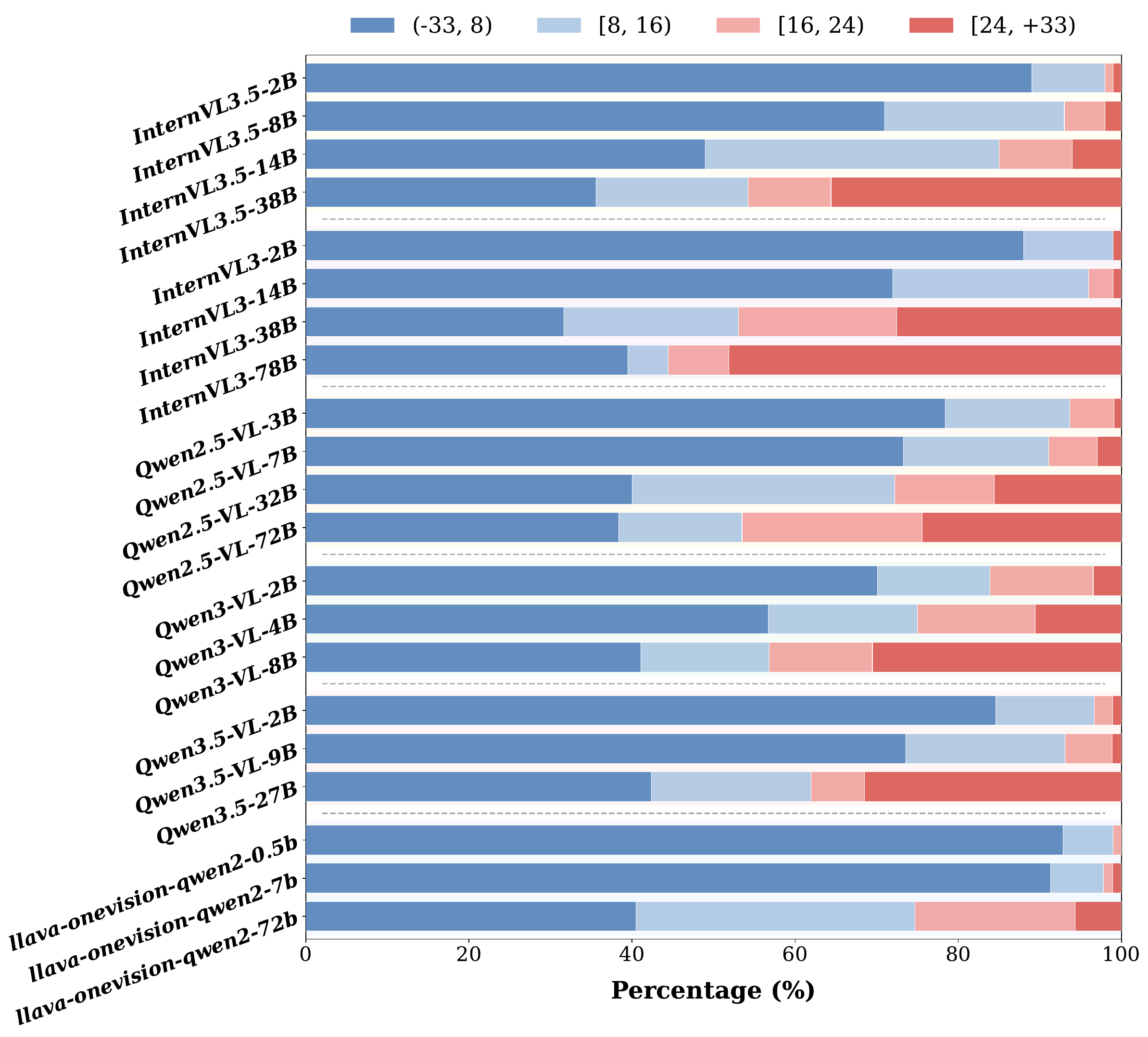} 
\caption{VIG distribution across models.}
\label{fig:vig_distribution}
\end{subfigure}
\hfill
\begin{subfigure}{0.48\textwidth}
\centering
\includegraphics[width=\linewidth]{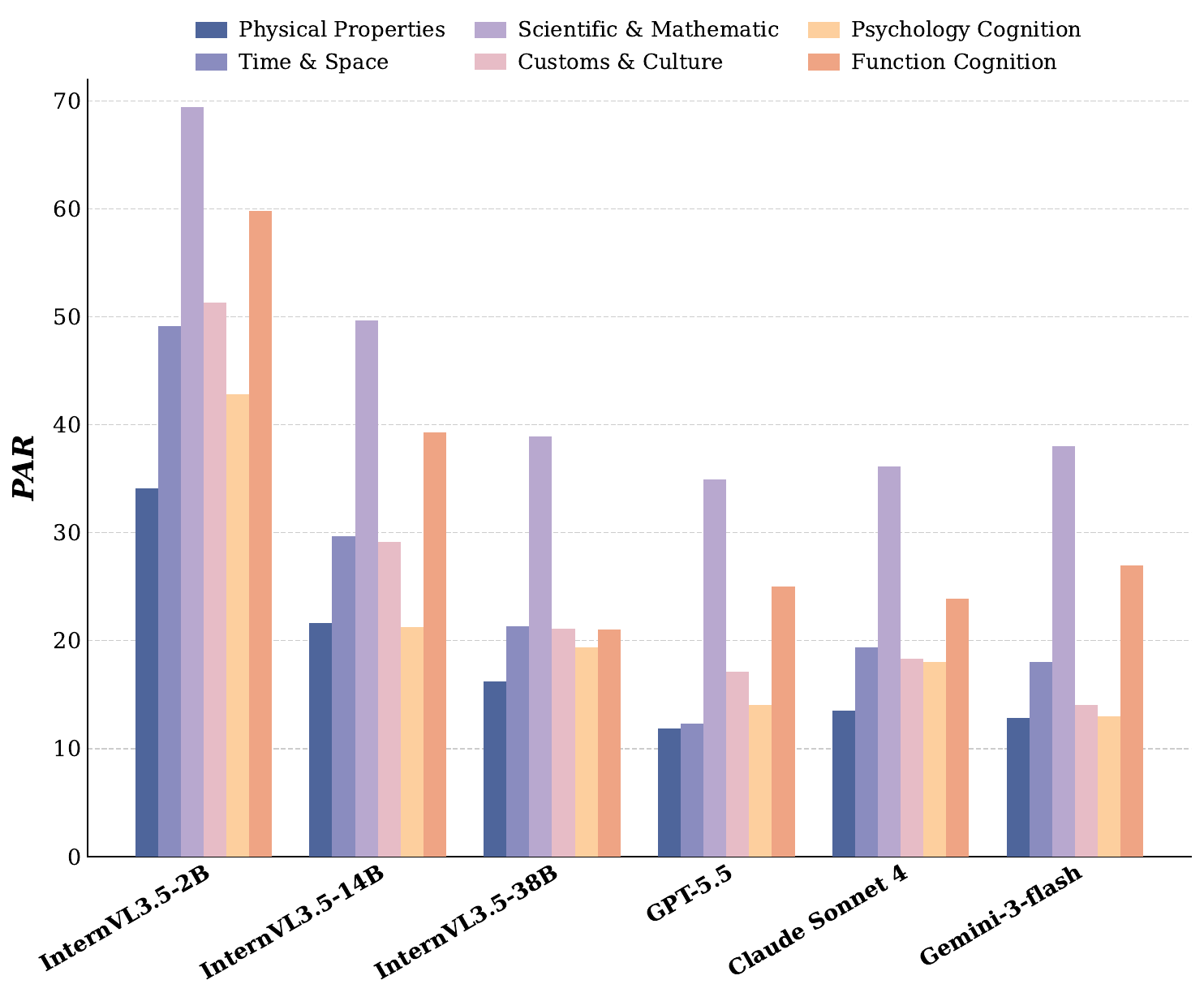} 
\caption{Per-category PAR across representative models.}
\label{fig:category_results}
\end{subfigure}
\caption{Overall analysis of model performance on ViSiP-Bench. Larger models exhibit a VIG polarization effect (a), and category-level prior resistance remains consistent across models (b).}
\label{fig:results-pic}
\end{figure*}


\subsection{Evaluation}
\label{sec:evaluation}

We evaluate 24 models on ViSiP-Bench, spanning six open-source model families (InternVL3.5 \citep{internvl3_5}, InternVL3 \citep{internvl3}, Qwen2.5-VL \citep{qwen25vl}, Qwen3-VL \citep{qwen3VL}, Qwen3.5 \citep{qwen35}, and LLaVA-OneVision \citep{an2025llava}) and three proprietary models (GPT-5.5 \citep{gpt55}, Claude Sonnet 4.6 \citep{claude46sonnet}, and Gemini-3-flash-preview \citep{geminiflash3preview}). Each experiment is repeated three times and results are averaged. For VIG computation, $\epsilon$ is set to $10^{-7}$, yielding a theoretical range of $(-33, 33)$. Table~\ref{tab:main_results} and Figure~\ref{fig:results-pic} present the results. We highlight three findings.

\noindent\textbf{Finding 1: Prior adherence is pervasive and not resolved by scale.}
Prior adherence remains a widespread issue across all evaluated models. Within open-source families, scaling up model size does not reliably reduce PAR; in certain categories, larger models do not demonstrate a consistent advantage in prior resistance.

\noindent\textbf{Finding 2: Larger models exhibit a VIG polarization effect.}
As shown in Figure~\ref{fig:vig_distribution}, while larger models accumulate a higher proportion of samples in the high VIG range, the proportion in the low VIG range does not decrease accordingly and in some families even increases, indicating that they either respond strongly to visual evidence or firmly adhere to their language priors, which we term \textit{VIG Polarization}.

\noindent\textbf{Finding 3: Category-level prior resistance is an intrinsic property of the prior type.}
As shown in Figure~\ref{fig:category_results}, prior resistance varies substantially across the six domains, yet this pattern remains largely consistent across models. \textit{Scientific \& Mathematic} exhibits the highest PAR values across all models, while \textit{Physical Properties} proves most amenable to visual override. A detailed per-category analysis is provided in Appendix~\ref{app:per-category}.

\begin{figure}[t]
\centering
\includegraphics[width=\columnwidth]{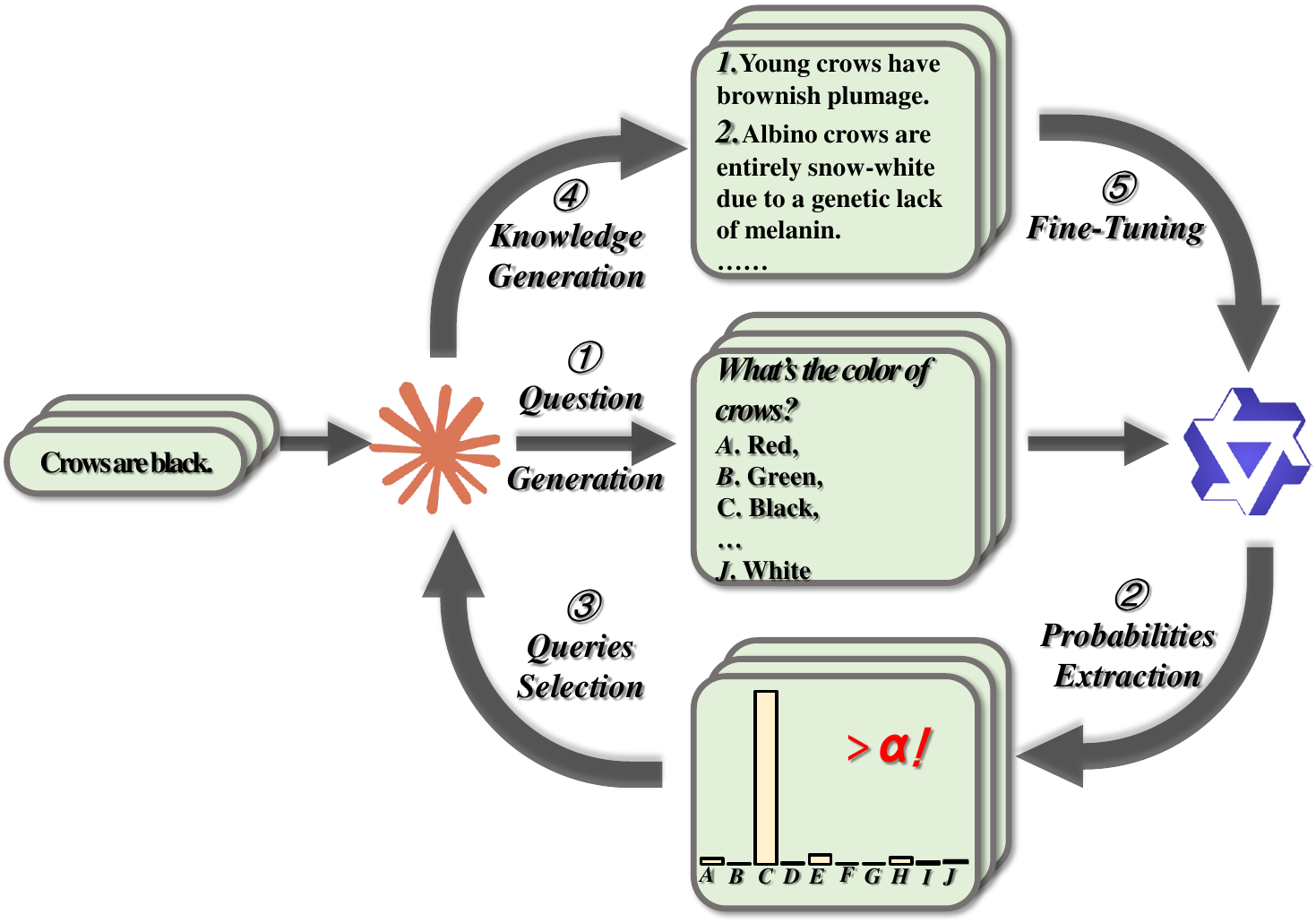}
\caption{Overview of the MAPS pipeline.}
\label{fig:MAPS}
\end{figure}

\section{Model-Aware Prior Softening }

\subsection{Motivation}

\S~\ref{sec:evaluation} demonstrates that language prior dependency persistently impairs visual perception across mainstream MLLMs and does not diminish with scale. Existing inference-time interventions offer a training-free alternative, but their effect is fundamentally bounded by what decoding-time signals can recover from a fixed network, making it difficult to achieve substantial prior resistance, as empirically confirmed in \S~\ref{sec:main_results}. Training-based methods offer stronger corrections, but existing approaches directly suppress the probability of prior-driven outputs during training, indiscriminately reducing the model's confidence in its prior knowledge regardless of whether that knowledge is harmful or benign. This risks disrupting the model's broader understanding of the real world and degrading general capabilities as a consequence.

We propose MAPS (Model-Aware Prior Softening), which takes a different stance. Rather than suppressing priors, MAPS softens them by enriching the model's knowledge with rare-but-real scenarios that the prior rarely accounts for. This leaves the model's existing knowledge intact while specifically targeting the priors that demonstrably interfere with visual perception.

\subsection{Method}

As illustrated in Figure~\ref{fig:MAPS}, MAPS operates in five steps.

\paragraph{Step 1: Question Generation.}
For each prior, Claude Opus 4.7 generates a question with 10 plausible answer options designed to elicit the model's prior belief.

\begin{table*}[t]
\centering
\scriptsize
\renewcommand{\arraystretch}{1.1}
\begin{tabular}{l
>{\centering\arraybackslash}p{1.1cm}
>{\centering\arraybackslash}p{1.3cm}
>{\centering\arraybackslash}p{0.5cm} |
>{\centering\arraybackslash}p{1.3cm}
>{\centering\arraybackslash}p{1.4cm}}
\toprule
\multicolumn{1}{c}{\textbf{Method}} & \textbf{PAR} $\downarrow$ & \textbf{Acc} $\uparrow$ & FC & $\mathbf{VIG_{avg}}$ $\uparrow$ & $\mathbf{VIG_{med}}$ $\uparrow$ \\
\midrule
\quad Base & 42.67 & 51.29 & 147 & 10.64 & 10.82\\
\midrule
\multicolumn{6}{c}{\textit{Inference-time Methods}} \\
\midrule
\quad VFP & 40.60 & 53.01 & 147 & 10.78 & 10.90\\
\quad CoT & 47.71& 43.11 & 147 & 8.11 & 8.32\\
\quad AS & 42.11 & 52.08 & 147 & 10.88& 11.22\\
\quad VCD & \underline{39.15} & \textbf{55.48} & 147 & \underline{12.11} & \underline{13.08} \\
\quad PAI & \textbf{39.01} & \underline{55.23} & 147 & \textbf{12.39} & \textbf{13.97}\\
\midrule
\multicolumn{6}{c}{\textit{Training-based Methods}} \\
\midrule
\quad Post-hoc & \textbf{32.05} & \textbf{61.94} & 385 & \textbf{19.81} & \textbf{18.92}\\
\quad NaPO & 37.09 & 57.34 & 209 & 16.33 & 17.12 \\
\quad \textbf{MAPS (Ours)} & \underline{35.09} & \underline{59.01} & 154 & \underline{18.84} & \underline{18.90} \\
\bottomrule
\end{tabular}
\caption{Comparison of inference-time and training-based methods on ViSiP-Bench evaluated on Qwen3-VL-8B-Instruct.}
\label{tab:intervention}
\end{table*}

\paragraph{Step 2: Probabilities Extraction.}
Each target model is queried on all generated questions, and the probability distribution over the 10 options is obtained via softmax over the output logits.

\paragraph{Step 3: Queries Selection.}
Questions where any single option receives a probability exceeding $\alpha$ (a hyperparameter controlling the strength of prior selection) are retained. These correspond to priors the model holds with high confidence, and their associated prior statements are selected.
\paragraph{Step 4: Knowledge Generation.}
For each selected prior statement, Claude Opus 4.7 generates 20 rare-but-real single-sentence descriptions that are factually grounded yet contradict the prior. To prevent data contamination, the generated content is kept entirely disjoint from ViSiP-Bench at the prompt level.

\paragraph{Step 5: Fine-Tuning.}
The rare-but-real sentences are aggregated into a model-specific corpus for supervised fine-tuning via LoRA, where each sentence serves as a training target to enrich the model's world knowledge.

\begin{table*}[t]
\centering
\scriptsize
\renewcommand{\arraystretch}{1.1}
\begin{tabular}{l
>{\centering\arraybackslash}p{1.6cm}
>{\centering\arraybackslash}p{1.3cm}
>{\centering\arraybackslash}p{1.3cm}
>{\centering\arraybackslash}p{1.7cm}}
\toprule
\textbf{Method} & \multicolumn{2}{c}{\textbf{Reasoning}} & \multicolumn{2}{c}{\textbf{General}} \\
\cmidrule(lr){2-3} \cmidrule(lr){4-5}
& \textbf{MMMU-Pro} $\uparrow$ & \textbf{VisuLogic} $\uparrow$ & \textbf{MMStar} $\uparrow$ & \textbf{RealWorldQA} $\uparrow$ \\
\midrule
\quad Base & \underline{55.90} & \textbf{28.14} & \textbf{70.90} & \underline{71.50}\\
\quad Post-hoc & 41.88 & 21.50 & 56.18 & 53.08\\
\quad NaPO & 49.51 & 24.93 & 66.51 & 66.00\\
\quad \textbf{MAPS (Ours)} & \textbf{56.84} & \underline{27.94} & \underline{69.27} & \textbf{71.68} \\
\bottomrule
\end{tabular}
\caption{Generalization results on reasoning and general vision benchmarks. All methods are evaluated on Qwen3-VL-8B-Instruct.}
\label{tab:generalization}
\end{table*}

\begin{table*}[t]
\centering
\scriptsize
\renewcommand{\arraystretch}{1.1}
\begin{tabular}{l
>{\centering\arraybackslash}p{1.1cm}
>{\centering\arraybackslash}p{1.1cm}
>{\centering\arraybackslash}p{0.5cm} |
>{\centering\arraybackslash}p{1.3cm}
>{\centering\arraybackslash}p{1.4cm} |
>{\centering\arraybackslash}p{1.7cm}
>{\centering\arraybackslash}p{1.1cm}}
\toprule
\textbf{Threshold} & \textbf{PAR} $\downarrow$ & \textbf{Acc} $\uparrow$ & FC & $\mathbf{VIG_{avg}}$ $\uparrow$ & $\mathbf{VIG_{med}}$ $\uparrow$ & \textbf{MMMU-Pro} $\uparrow$ & \textbf{MMStar} $\uparrow$ \\
\midrule
$\alpha = 0.50$ & 33.09 & 60.93 & 218 & 19.81 & 19.08 & 50.90 & 63.84\\
$\alpha = 0.80$ & 34.24 & 60.10 & 181 & 19.41 & 19.90 & 54.17 & 66.81\\
$\alpha = 0.90$ & 35.09 & 59.01 & 154 & 18.84 & 18.90 & 56.84 & 69.27\\
$\alpha = 0.95$ & 38.41 & 54.09 & 150 & 17.77 & 18.13 & 56.10 & 71.02\\
$\alpha = 1.00$ & 42.67 & 51.29 & 147 & 10.64 & 10.82 & 55.90 & 70.90 \\
\bottomrule
\end{tabular}
\caption{Ablation study on the prior probing threshold $\alpha$. Results on both ViSiP-Bench and generalization benchmarks are reported. $\alpha = 1.00$ corresponds to the base model without any fine-tuning.}
\label{tab:ablation}
\end{table*}

\section{Experiments}

\subsection{Experimental Setup}


\noindent\textbf{Implementation Details.}
All training-based methods are implemented on 4 NVIDIA A100 80GB GPUs. MAPS is applied to Qwen3-VL-8B-Instruct via LoRA fine-tuning for 2 epochs, with the prior probing threshold set to $\alpha = 0.90$ unless specified. To assess whether these improvements generalize beyond a single backbone, we further apply MAPS under an identical configuration to six additional models spanning three architecture families, with full results reported in Appendix~\ref{app:crossmodel}.

\noindent\textbf{Baselines.}
We compare MAPS against representative methods from both inference-time and training-based families. 

For inference-time methods:\textit{VFP} (Visual-Focused Prompt) instructs the model to focus on the image before responding; \textit{CoT}~\citep{cot} prompts step-by-step reasoning; \textit{AS} (Attention Steering) uniformly amplifies visual token attention weights without modifying logits; \textit{VCD}~\citep{VCD} subtracts logits from a perturbed-image forward pass to suppress prior-driven tokens; \textit{PAI}~\citep{PAI} combines visual attention amplification with text-only logit subtraction as a language prior basal. 

For training-based methods:
\textit{Post-hoc}~\citep{Post-hoc} penalizes prior-driven outputs via empty-image probability estimation; \textit{NaPO}~\citep{NaPO} applies noise-aware preference optimization to suppress the model's preference for language-prior-driven answers over visually grounded ones.

\noindent\textbf{Benchmarks for Generalization.}
We evaluate generalization on two dimensions. For visual reasoning, we use MMMU-Pro~\citep{mmmu-pro}, which explicitly filters out text-only solvable questions to assess genuine visual reasoning across multiple disciplines, and VisuLogic~\citep{VisuLogic}, which eliminates all textual shortcuts and focuses purely on vision-based logical deduction. For general visual understanding, we use MMStar~\citep{mmstar}, which covers six core visual capabilities, and RealWorldQA, which tests spatial and physical understanding on authentic real-world photographs.

\subsection{Main Results}
\label{sec:main_results}

Table~\ref{tab:intervention} presents the results of all methods on ViSiP-Bench. Inference-time methods yield limited improvements overall, while training-based methods prove more effective but differ substantially in how they achieve gains.

Among inference-time methods, CoT consistently hurts performance, confirming that extended reasoning amplifies prior adherence rather than reducing it. VFP and AS yield only marginal gains, suggesting that neither explicit visual instruction nor attention redistribution is sufficient to shift the model's confidence toward visual evidence. VCD and PAI achieve the best inference-time results, indicating that directly suppressing prior-driven logits at the decoding stage is a more effective direction, though the overall improvement remains limited.

Among training-based methods, Post-hoc achieves the lowest PAR but at a significant cost: FC increases from 147 to 385, indicating that its probability penalization severely disrupts the model's prior knowledge and causes prior responses to collapse toward random. NaPO improves over the baseline but with limited gains. MAPS achieves the strongest PAR and VIG improvements while maintaining FC at 154, close to the baseline. This demonstrates that supplementing the model with factually grounded rare-but-real knowledge, rather than suppressing priors directly, is a more effective and less destructive approach to enhancing visual reliance.

This advantage is not specific to Qwen3-VL-8B-Instruct. Extending MAPS to six additional models spanning three architecture families yields consistent reductions in PAR and gains in VIG, with FC remaining close to the base model throughout, as reported in Table~\ref{tab:crossmodel_visip} of Appendix~\ref{app:crossmodel}.

\subsection{Further Analysis}

\noindent\textbf{Generalization Analysis.}
Table~\ref{tab:generalization} reports results on reasoning and general vision benchmarks. Post-hoc suffers substantial degradation across all four benchmarks, consistent with its destructive effect on prior knowledge observed in Table~\ref{tab:intervention}. NaPO also declines across all tasks, though to a lesser extent, indicating that preference-based optimization incurs a non-trivial cost to general capabilities. MAPS maintains performance close to the baseline across all benchmarks, demonstrating that it leaves the model's reasoning and general visual capabilities largely intact. Unlike prior suppression methods that sacrifice general capabilities for prior resistance, MAPS acquires more comprehensive world knowledge through rare-but-real supervision, preserving the model's understanding of the real world.

This preservation effect likewise generalizes across models and architectures. Extending MAPS to six additional models leaves performance on all four generalization benchmarks within a narrow margin, with the largest absolute change reaching only 0.87 points and no consistent direction of degradation, as reported in Table~\ref{tab:crossmodel_general} of Appendix~\ref{app:crossmodel}.

\noindent\textbf{Ablation on Prior Probing Threshold.}
To isolate the effect of the threshold $\alpha$ from training data quantity, we control the total training data to match that of $\alpha=0.90$. For lower thresholds where more priors are selected, we proportionally reduce the number of rare-but-real descriptions per prior; for higher thresholds where fewer priors are selected, we prompt Claude Opus 4.7 to generate additional descriptions per prior to reach the same total. Table~\ref{tab:ablation} reports results across four threshold values. A low threshold such as $\alpha=0.50$ selects a large number of priors indiscriminately, including many that the model does not strongly hold, which interferes with targeted prior softening and leads to suboptimal performance on both ViSiP-Bench and generalization benchmarks. A high threshold such as $\alpha=0.95$ is overly restrictive, covering too few priors to achieve meaningful improvement in visual perception. $\alpha=0.90$ achieves the best balance, maximally improving visual perception while preserving the model's reasoning and general capabilities.

\section{Conclusion}


We present ViSiP-Bench, a redesigned benchmark for evaluating language prior dependency in MLLMs. Existing benchmarks rely on a shared-prior assumption, yet cross-model prior agreement falls below 80\%, and accuracy overestimates prior resistance as over 15\% of items can be answered without any image, motivating a re-examination of how prior resistance should be measured. 

ViSiP-Bench addresses these issues through a model-aware dynamic prior induction protocol paired with three complementary metrics. PAR and Filtered Accuracy capture prior resistance at the decision level by conditioning on each model's own elicited prior, while VIG provides a continuous measure of the confidence shift induced by visual evidence. Evaluation of 24 models reveals that prior adherence is pervasive across model families, does not diminish with scale, and manifests as VIG Polarization in larger models, which either override their priors decisively or adhere to them just as firmly. These findings suggest that inference-time interventions alone are insufficient, motivating a training-based solution that avoids indiscriminately suppressing the model's world knowledge. 

We therefore propose MAPS, which identifies each model's most strongly held priors and softens them by enriching the model with targeted, factually grounded knowledge of rare-but-real scenarios, rather than suppressing priors indiscriminately. MAPS achieves substantial improvements in PAR and VIG on ViSiP-Bench while preserving performance on reasoning and general vision benchmarks, and this behavior holds consistently across seven models spanning three architecture families, demonstrating that enhanced visual reliance can be obtained without sacrificing a model's broader capabilities.



\section*{Limitations}

First, the six categories in ViSiP-Bench, though designed to broadly cover commonsense knowledge relevant to MLLMs, follow a fixed taxonomy and cannot capture every form of world knowledge a model may rely on. Priors rooted in specialized domains or in rapidly evolving knowledge fall outside this taxonomy, and whether the patterns of prior adherence we observe extend to such cases remains an open question for future benchmarks to address.

Second, VIG relies on token-level output probabilities and is therefore inapplicable to closed models accessed only through APIs. This leaves the VIG Polarization phenomenon reported in this work unverified for proprietary models, which may exhibit different confidence dynamics than what we observe here.

\section*{Ethical Considerations}
Data construction relies on proprietary generative models, including Claude Opus 4.7 for prior enumeration and scenario generation, and GPT-Image-2 and Nano Banana 2 for image synthesis, all accessed through official APIs in accordance with the respective providers' usage policies. Evaluation additionally involves querying proprietary models such as GPT-5.5, Claude Sonnet 4.6, and Gemini-3-flash-preview through their official APIs, again in compliance with the applicable terms of service. Content synthesized by generative models, as well as web-sourced images, is subject to the same three-criteria quality review, which additionally serves to filter out any inappropriate or misleading generations before inclusion in ViSiP-Bench. Human annotation involved in question and option design, as well as quality verification, was carried out by voluntary annotators who were fully informed of the task and compensated fairly for their time; the annotation task does not involve exposure to violent, explicit, or otherwise harmful content, as all images depict ordinary, if uncommon, real-world scenarios.

We note that generative AI tools were employed only for editing purposes, such as improving wording and sentence clarity, and played no role in the underlying research. All ideas, experiments, analyses, and conclusions presented in this paper originate from and were validated by the authors.

\section*{Acknowledgments}
This work was supported by the
National Natural Science Foundation of China
(Grant No.U24A20250), Sichuan Science and
Technology Program (No.2025ZNSFSC1487), the Fundamental
Research Funds for the Central Universities
(No.ZYGX2024J022 and No.ZYGX2024Z005),
and the Science, Technology and Innovation
Project of Shenzhen Longhua District (No.20260309G23410662).


\bibliography{custom}

\appendix

\section{Data Quality Analysis}
\label{app:dataqualityAnalysis}

\subsection{Identification of No-Image Correct Responses}

Over 15\% of items in existing benchmarks can be answered correctly without any image. To identify the source of each such response and categorize it into one of three types (\textit{Chance Success}, \textit{Low-quality Framing}, or \textit{Anomalous Prior}), we adopt a hybrid pipeline combining automated screening with human verification.

The pipeline proceeds as follows. All items answered correctly under the Blank condition are first collected as candidates. Each candidate then undergoes the following three-stage procedure.

\begin{table*}[t]
\centering
\small
\renewcommand{\arraystretch}{1.3}
\begin{tabular}{
>{\raggedright\arraybackslash}p{3cm}
>{\raggedright\arraybackslash}p{2cm}
>{\raggedright\arraybackslash}p{4cm}
>{\raggedright\arraybackslash}p{4cm}}
\toprule
\textbf{Source} & \textbf{Model} & \textbf{Question} & \textbf{Reason for Categorization} \\
\midrule
\textit{Chance Success} & \multicolumn{1}{c}{Qwen3-VL-2B} & When is the vernal equinox? (A. March 20 B. June 20 C. July 23 D. December 21) & Model assigns near-uniform probabilities: A: 28.3\%, B: 24.1\%, C: 26.8\%, D: 20.8\%, indicating no strong prior and random correct selection. \\
\midrule
\textit{Low-quality Framing} & \multicolumn{1}{c}{---} & What is unusual about the color of the rose in this image? (A. Red B. Black C. Pink D. White) & The phrase ``unusual'' implicitly hints at an atypical answer, guiding the model toward anomalous options. \\
\midrule
\textit{Low-quality Framing} & \multicolumn{1}{c}{---} & What color is the watermelon? (A. Red B. Green C. White D. Yellow) & The question does not specify flesh or rind, making the prior ambiguous and the correct answer uncertain without the image. \\
\midrule
\textit{Anomalous Prior} & \multicolumn{1}{c}{Qwen2.5-VL-3B} & What color is the dolphin? (A. Blue B. White C. Pink D. Grey) & Model assigns high probability to ``Pink'', indicating it inherently holds the anomalous answer as its own prior. \\
\bottomrule
\end{tabular}
\caption{Representative examples of the three sources of no-image correct responses.}
\label{tab:appendix_cases}
\end{table*}
\noindent\textbf{Stage 1: Human Verification of Question Quality.}
Human annotators examine each candidate question along two dimensions. First, they assess whether the question is associated with a sufficiently strong and unambiguous commonsense prior. Questions where the correct answer is genuinely uncertain even without an image (for example, ``what color is the dragon fruit?'' where both red and white are valid) are flagged as \textit{Low-quality Framing} due to an insufficiently strong prior. Second, annotators check whether the question wording contains implicit hints that guide the model toward the anomalous answer. For instance, a question such as ``what is unusual about the color of the rose in this image?'' implicitly signals that the rose's color is atypical, nudging the model toward selecting an uncommon option such as black or white. Such items are also categorized as \textit{Low-quality Framing}.

\noindent\textbf{Stage 2: Automated Probability Analysis.}
Items that pass Stage 1 (i.e., those with a strong and unambiguous prior and no implicit hints) are further analyzed by re-querying the target model under the Blank condition and extracting the full probability distribution over all answer options via softmax over the output logits. If the variance across option probabilities is low, indicating that the model assigns similar probabilities to all options, the item is categorized as \textit{Chance Success}. This pattern is particularly prevalent in smaller models with weaker prior knowledge, where correct responses arise from random selection rather than genuine prior activation.
In our implementation, if the maximum probability among all options falls below 65\%, the item is categorized as \textit{Chance Success}.

\noindent\textbf{Stage 3: Anomalous Prior Identification.}
If the probability distribution exhibits high variance, with one option receiving substantially higher probability than the others, the model is judged to hold a strong prior for this question. Since the model answers correctly under the Blank condition despite the item being designed to conflict with commonsense expectations, this indicates that the model inherently treats the anomalous answer as its own prior. Such items are categorized as \textit{Anomalous Prior}.

\subsection{Typical Cases}

Table~\ref{tab:appendix_cases} provides representative examples of each source type identified by the above pipeline.

\section{Visual Style Examples}
\label{app:VisualStyleExamples}

\begin{figure*}[t]
\centering
\begin{subfigure}{0.48\textwidth}
\centering
\includegraphics[width=\linewidth]{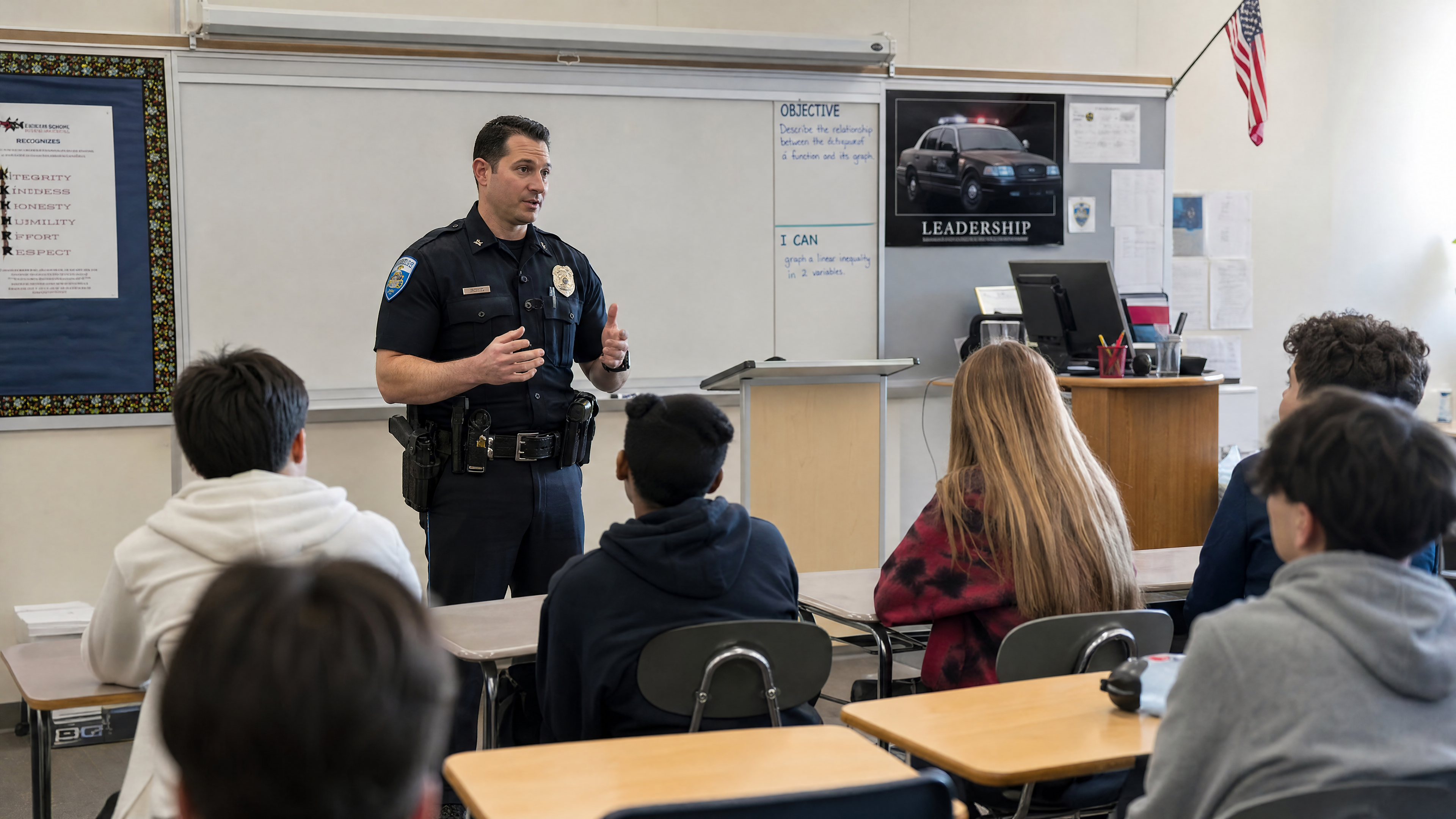}
\caption{Photorealistic: A police officer shares career experiences in a classroom.}
\label{fig:style_photo}
\end{subfigure}
\hfill
\begin{subfigure}{0.48\textwidth}
\centering
\includegraphics[width=\linewidth]{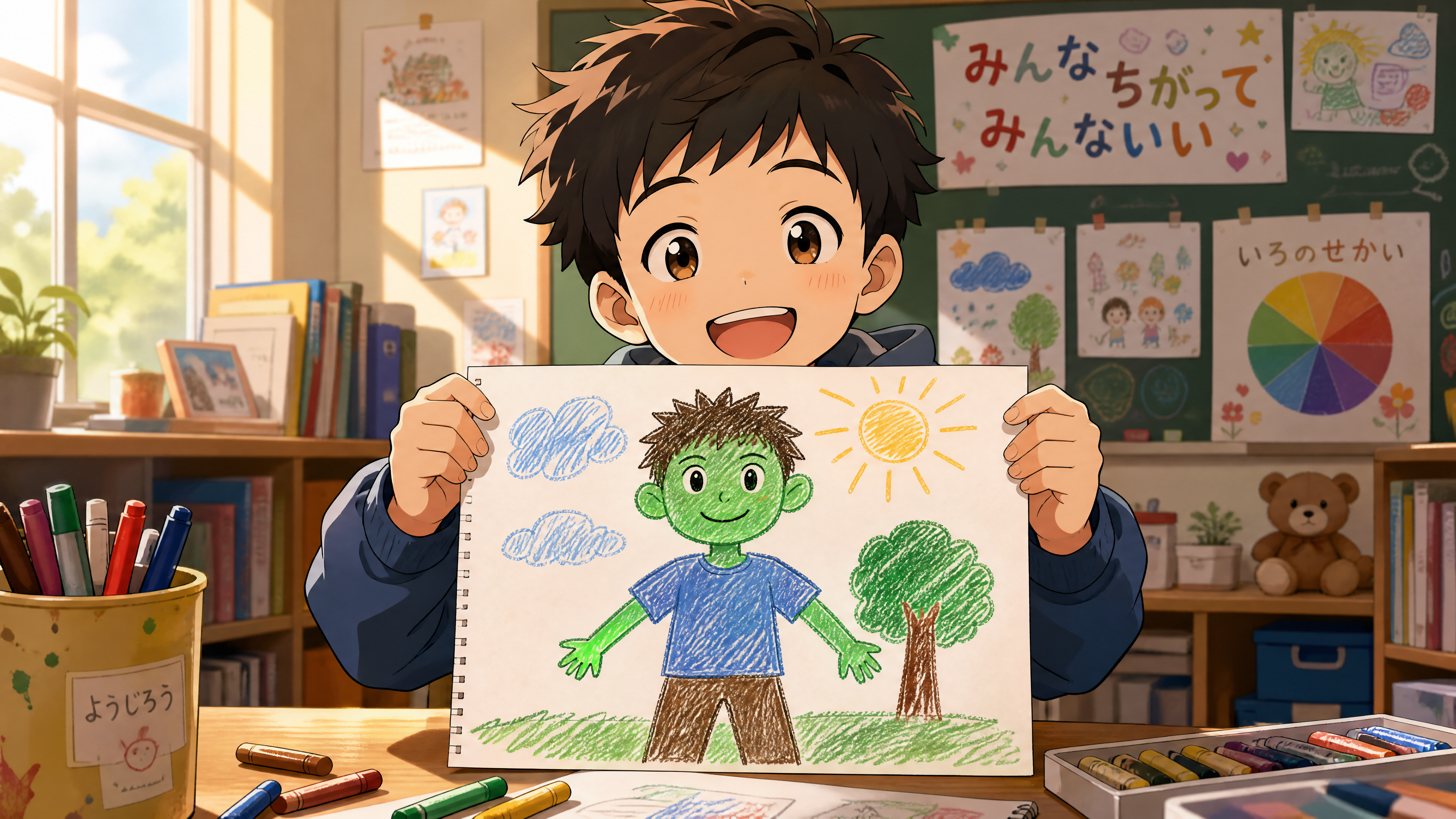}
\caption{Anime: A color-blind boy proudly displays his drawing of a green-skinned figure.}
\label{fig:style_anime}
\end{subfigure}
\vspace{1em}
\begin{subfigure}{0.48\textwidth}
\centering
\includegraphics[width=\linewidth]{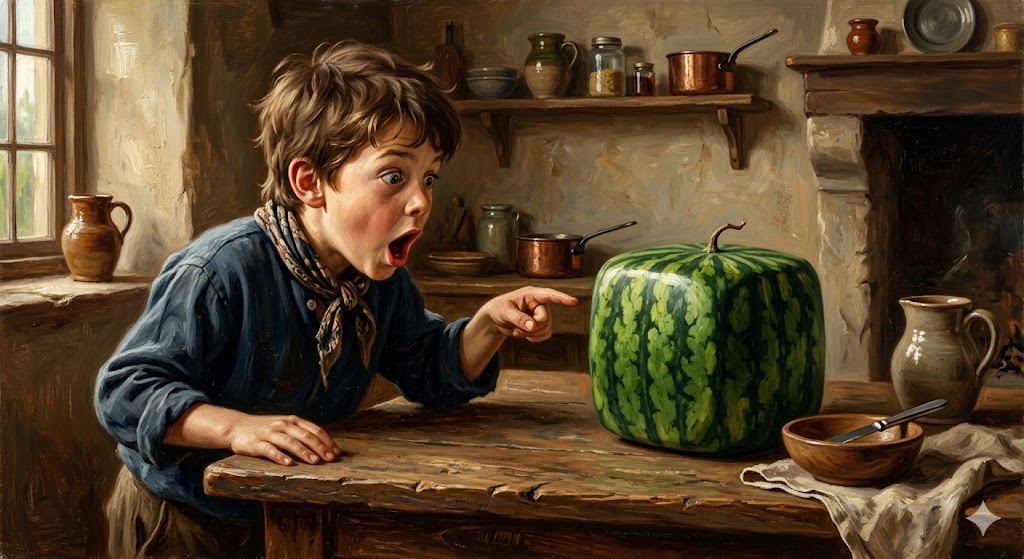}
\caption{Oil painting: A boy stares in disbelief at a square watermelon.}
\label{fig:style_oil}
\end{subfigure}
\hfill
\begin{subfigure}{0.48\textwidth}
\centering
\includegraphics[width=\linewidth]{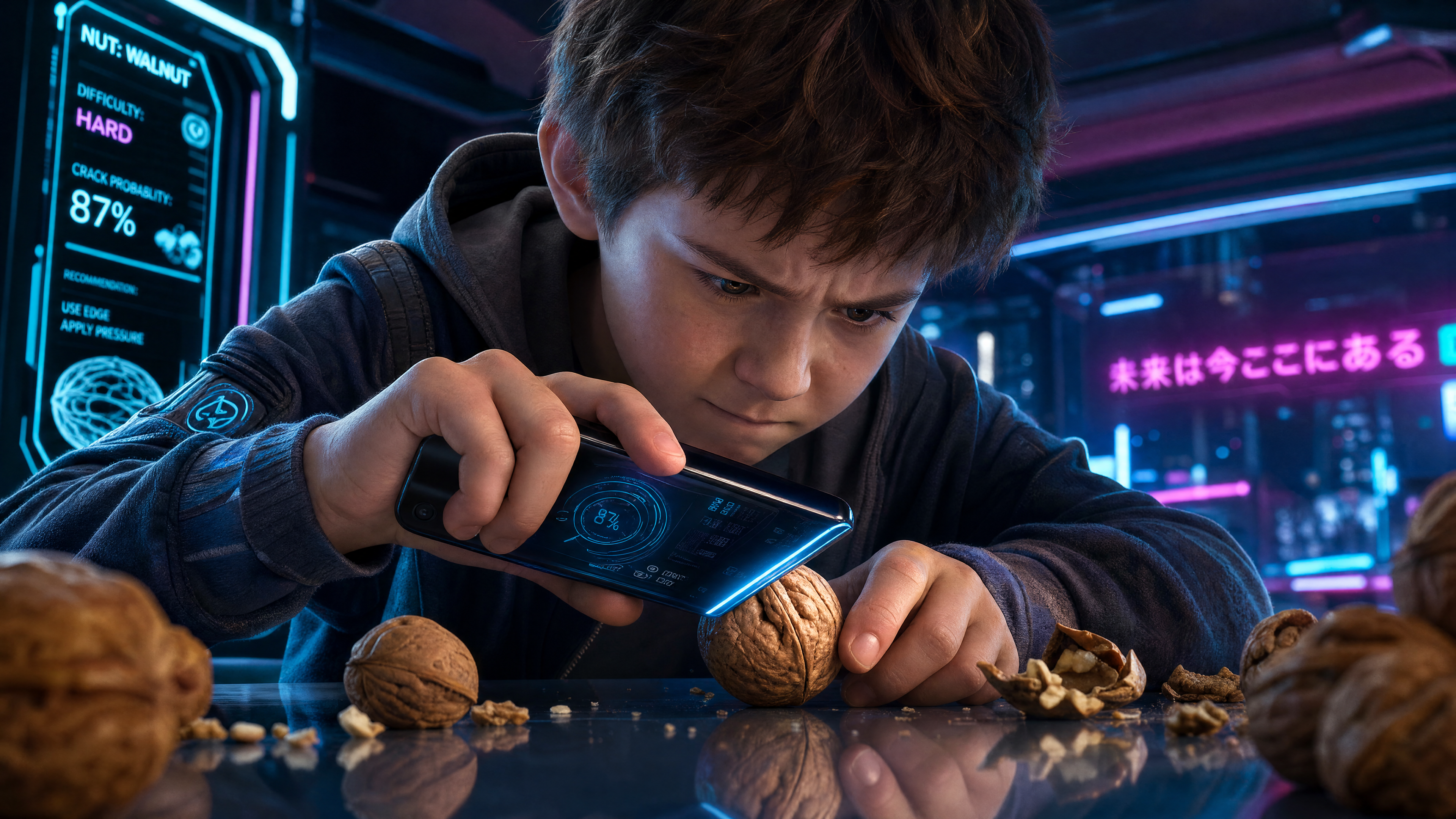}
\caption{Sci-fi: A boy cracks a walnut using his smartphone.}
\label{fig:style_scifi}
\end{subfigure}
\caption{Representative examples of the four visual styles in ViSiP-Bench. Each image depicts a rare-but-real scenario designed to conflict with commonsense priors.}
\label{fig:style_examples}
\end{figure*}

ViSiP-Bench incorporates images across four distinct visual styles to ensure that model performance is not confounded by image style or source. Table~\ref{tab:style_prompts} lists the generation prompts used for each style, and Figure~\ref{fig:style_examples} presents one representative example from each style.

\begin{table}[h]
\centering
\small
\renewcommand{\arraystretch}{1.3}
\begin{tabular}{>{\raggedright\arraybackslash}p{2cm} >{\raggedright\arraybackslash}p{4.5cm}}
\toprule
\textbf{Style} & \textbf{Generation Prompt} \\
\midrule
Photorealistic & A police officer in full uniform stands at the front of a classroom, sharing career experiences with attentive students seated at desks. \\
\midrule
Anime & A color-blind boy proudly holds up his drawing of a figure with green skin, smiling with satisfaction, in a vibrant anime art style. \\
\midrule
Oil painting & A young boy stares in disbelief at a perfectly square watermelon sitting on a wooden table, his mouth wide open in shock, rendered in the style of a classical oil painting. \\
\midrule
Sci-fi & A boy uses his smartphone to crack open a walnut, pressing the screen edge against the shell with focused determination, in a futuristic sci-fi setting with neon lighting. \\
\bottomrule
\end{tabular}
\caption{Generation prompts for each visual style in ViSiP-Bench.}
\label{tab:style_prompts}
\end{table}

\section{Cross-Model and Cross-Family Validation of MAPS}
\label{app:crossmodel}
To verify that the effectiveness of MAPS is not specific to a single backbone, we apply MAPS to six additional models spanning three architecture families: InternVL3.5 (2B, 8B), InternVL3 (14B), Qwen2.5-VL (7B), Qwen3.5 (9B), and LLaVA-OneVision (7B). All models are fine-tuned under an identical configuration (prior probing threshold $\alpha = 0.90$, LoRA fine-tuning for 2 epochs), with no per-model hyperparameter adjustment, to verify that the improvement generalizes without model-specific tuning.

\begin{table*}[h]
\centering
\scriptsize
\renewcommand{\arraystretch}{1.1}
\begin{tabular}{l l
>{\centering\arraybackslash}p{0.85cm}
>{\centering\arraybackslash}p{0.75cm}
>{\centering\arraybackslash}p{0.4cm}
>{\centering\arraybackslash}p{0.95cm}
>{\centering\arraybackslash}p{0.95cm}}
\toprule
\textbf{Model} & \textbf{Type} & \textbf{PAR}$\downarrow$ & \textbf{Acc}$\uparrow$ & \textbf{FC} & $\mathbf{VIG_{avg}}$$\uparrow$ & $\mathbf{VIG_{med}}$$\uparrow$ \\
\midrule
\multirow{2}{*}{InternVL3.5-2B} & base & 51.51 & 42.94 & 182 & 4.41 & 4.11 \\
 & +MAPS & \textbf{46.91} & \textbf{47.11} & 189 & \textbf{8.91} & \textbf{9.20} \\
\midrule
\multirow{2}{*}{InternVL3.5-8B} & base & 47.17 & 44.64 & 156 & 7.13 & 7.48 \\
 & +MAPS & \textbf{39.11} & \textbf{51.85} & 158 & \textbf{15.94} & \textbf{15.63} \\
\midrule
\multirow{2}{*}{InternVL3-14B} & base & 48.40 & 41.33 & 161 & 6.21 & 6.33 \\
 & +MAPS & \textbf{42.60} & \textbf{47.91} & 166 & \textbf{13.81} & \textbf{13.48} \\
\midrule
\multirow{2}{*}{LLaVA-OV-7B} & base & 67.10 & 25.36 & 317 & 3.39 & 2.48 \\
 & +MAPS & \textbf{63.01} & \textbf{29.81} & 319 & \textbf{5.94} & \textbf{6.11} \\
\midrule
\multirow{2}{*}{Qwen2.5-VL-7B} & base & 55.35 & 34.51 & 198 & 5.56 & 4.93 \\
 & +MAPS & \textbf{51.91} & \textbf{38.13} & 201 & \textbf{10.63} & \textbf{11.21} \\
\midrule
\multirow{2}{*}{Qwen3.5-9B} & base & 51.24 & 39.36 & 137 & 5.24 & 5.24 \\
 & +MAPS & \textbf{43.22} & \textbf{47.51} & 141 & \textbf{12.97} & \textbf{13.62} \\
\bottomrule
\end{tabular}
\caption{Cross-model results on ViSiP-Bench. MAPS consistently reduces PAR and increases Filtered Accuracy and VIG across all six models spanning three architecture families, while FC remains close to the base model, indicating that prior knowledge is not indiscriminately disrupted.}
\label{tab:crossmodel_visip}
\end{table*}

Table~\ref{tab:crossmodel_visip} reports results on ViSiP-Bench. Across all six models, MAPS consistently reduces PAR (by 3.44 to 8.06 points) and improves Filtered Accuracy (by 3.62 to 8.15 points), with no exceptions across model scale (2B--14B) or family. FC increases only marginally (by 2 to 7 samples), in sharp contrast to the collapse observed for Post-hoc in Table~\ref{tab:intervention} (147 $\to$ 385), confirming that MAPS softens rather than indiscriminately erodes prior knowledge. VIG improves substantially for every model, in most cases more than doubling, indicating that the effect is not confined to a single backbone but reflects a general property of the model-aware prior softening procedure. We note that the absolute PAR reduction for LLaVA-OneVision-7B (-4.09) is somewhat smaller than for other models; this model also exhibits the highest base PAR (67.10) among the six, suggesting that models with a very large number of strongly held priors may require a longer fine-tuning schedule or a larger rare-but-real corpus to achieve proportionally similar gains.

\begin{table*}

\centering
\scriptsize
\renewcommand{\arraystretch}{1.1}
\begin{tabular}{l l
>{\centering\arraybackslash}p{1.0cm}
>{\centering\arraybackslash}p{1.0cm}
>{\centering\arraybackslash}p{0.9cm}
>{\centering\arraybackslash}p{1.2cm}}
\toprule
\textbf{Model} & \textbf{Type} & \textbf{MMMU-Pro} & \textbf{VisuLogic} & \textbf{MMStar} & \textbf{RealWorldQA} \\
\midrule
\multirow{2}{*}{InternVL3.5-2B} & base & 34.10 & 24.10 & 61.18 & 63.12 \\
 & +MAPS & 33.91 & 24.15 & 62.05 & 62.49 \\
\midrule
\multirow{2}{*}{InternVL3.5-8B} & base & 41.20 & 28.60 & 65.23 & 67.32 \\
 & +MAPS & 41.13 & 28.71 & 64.79 & 67.51 \\
\midrule
\multirow{2}{*}{InternVL3-14B} & base & 40.10 & 26.30 & 64.14 & 63.98 \\
 & +MAPS & 40.23 & 26.41 & 64.87 & 63.48 \\
\midrule
\multirow{2}{*}{LLaVA-OV-7B} & base & 24.10 & 25.30 & 56.13 & 55.30 \\
 & +MAPS & 24.18 & 26.11 & 55.84 & 55.15 \\
\midrule
\multirow{2}{*}{Qwen2.5-VL-7B} & base & 38.30 & 26.00 & 59.14 & 61.38 \\
 & +MAPS & 38.19 & 25.63 & 59.71 & 60.91 \\
\midrule
\multirow{2}{*}{Qwen3.5-9B} & base & 42.10 & 28.61 & 66.51 & 68.31 \\
 & +MAPS & 42.34 & 27.90 & 66.90 & 68.44 \\
\bottomrule
\end{tabular}
\caption{Generalization results on reasoning (MMMU-Pro, VisuLogic) and general vision (MMStar, RealWorldQA) benchmarks across six models. MAPS preserves performance within a narrow margin for every model, with no consistent direction of degradation.}
\label{tab:crossmodel_general}
\end{table*}

Table~\ref{tab:crossmodel_general} reports generalization results on the same six models. The largest absolute change across all 24 model--benchmark pairs is 0.87 points (MMStar, InternVL3.5-2B), and changes are not consistently signed: some models show small gains on certain benchmarks and small losses on others, with no systematic degradation pattern. This indicates that the preservation of general reasoning and visual understanding capabilities observed for Qwen3-VL-8B-Instruct in Table~\ref{tab:generalization} is not an artifact of that particular model, but a consistent property of MAPS across scales and architecture families.

\section{Annotation Guidelines}
\label{app:annatation}

\noindent\textbf{Task Description.}
You will be presented with an image depicting a rare-but-real scenario that conflicts with commonsense expectations. Your task is to design a multiple-choice question that requires the respondent to carefully observe the image in order to answer correctly. The correct answer must be visually grounded in the image, and must differ clearly from what one would expect based on commonsense knowledge alone.

\noindent\textbf{What is a Commonsense Prior?}
A commonsense prior is a default expectation about the world that a person (or model) holds based on general knowledge. For example, most people expect swans to be white. If an image shows a black swan, a respondent relying on their prior rather than the image would incorrectly answer ``white''. Questions should be designed so that relying on the prior leads to a wrong answer, while carefully observing the image leads to the correct one.

\noindent\textbf{Question Design Requirements.}

(1) Identify the anomalous element in the image and formulate a question that directly targets it.

(2) Phrase the question in a neutral and objective manner. Do \textbf{not} use any wording that hints at the anomalous nature of the image (e.g., avoid phrases such as ``what is unusual about...'').

(3) Provide exactly four answer choices. One choice must correspond to the visually correct answer; the remaining three must be plausible distractors grounded in commonsense expectations.

(4) Ensure that no option is obviously implausible, as this would allow the correct answer to be identified by elimination.

\begin{table*}[t]
\centering
\small
\renewcommand{\arraystretch}{1.1}
\begin{tabular}{l
>{\centering\arraybackslash}p{1.3cm}
>{\centering\arraybackslash}p{1.3cm}
>{\centering\arraybackslash}p{1.3cm}
>{\centering\arraybackslash}p{1.3cm}
>{\centering\arraybackslash}p{1.3cm}
>{\centering\arraybackslash}p{1.3cm}}
\toprule
\textbf{Model} & \textbf{Physical} & \textbf{Time \&} \textbf{Space} & \textbf{Scientific} & \textbf{Customs \&} \textbf{Culture} & \textbf{Psychology} & \textbf{Function} \\
\midrule
\multicolumn{7}{c}{\textit{Open-source Models}} \\
\midrule
InternVL3.5-2B & 34.1 & 49.2 & 69.4 & 51.3 & 43.2 & 60.1 \\
InternVL3.5-8B & 30.2 & 44.1 & 63.2 & 47.1 & 39.4 & 55.2 \\
InternVL3.5-14B & 22.1 & 29.8 & 49.7 & 29.1 & 21.3 & 39.4 \\
InternVL3.5-38B & 16.3 & 21.4 & 38.9 & 21.4 & 19.7 & 21.1 \\
\midrule[0.04em]
InternVL3-2B & 40.2 & 55.3 & 78.4 & 58.2 & 51.3 & 68.2 \\
InternVL3-14B & 28.4 & 41.2 & 62.1 & 44.3 & 37.2 & 52.1 \\
InternVL3-38B & 20.1 & 31.4 & 52.3 & 33.2 & 28.4 & 41.2 \\
InternVL3-78B & 18.2 & 29.1 & 49.8 & 31.4 & 26.1 & 38.9 \\
\midrule[0.04em]
Qwen2.5-VL-3B & 41.3 & 56.2 & 79.1 & 59.4 & 52.1 & 69.3 \\
Qwen2.5-VL-7B & 33.2 & 48.1 & 68.3 & 50.2 & 42.3 & 59.1 \\
Qwen2.5-VL-32B & 24.3 & 36.2 & 58.4 & 38.1 & 31.2 & 47.3 \\
Qwen2.5-VL-72B & 19.4 & 30.1 & 51.2 & 32.3 & 27.4 & 40.1 \\
\midrule[0.04em]
Qwen3-VL-2B & 38.2 & 52.1 & 74.3 & 55.1 & 48.2 & 64.3 \\
Qwen3-VL-4B & 29.3 & 43.2 & 65.1 & 45.2 & 38.3 & 54.2 \\
Qwen3-VL-8B & 21.4 & 32.1 & 54.2 & 34.3 & 29.1 & 43.4 \\
\midrule[0.04em]
Qwen3.5-2B & 34.3 & 48.2 & 70.1 & 51.2 & 44.3 & 61.2 \\
Qwen3.5-9B & 28.2 & 42.1 & 64.3 & 44.1 & 37.2 & 53.3 \\
Qwen3.5-27B & 20.3 & 30.2 & 52.1 & 32.4 & 27.3 & 41.3 \\
\midrule[0.04em]
LLaVA-OV-0.5B & 48.2 & 63.1 & 84.3 & 65.2 & 58.1 & 74.2 \\
LLaVA-OV-7B & 43.1 & 58.2 & 79.4 & 60.3 & 53.2 & 69.1 \\
LLaVA-OV-72B & 36.2 & 51.3 & 73.2 & 53.4 & 46.3 & 62.4 \\
\midrule
\multicolumn{7}{c}{\textit{Proprietary Models}} \\
\midrule
GPT-5.5 & 12.1 & 11.9 & 35.2 & 15.4 & 14.2 & 25.1 \\
Claude Sonnet 4.6 & 13.4 & 19.4 & 36.1 & 18.3 & 19.1 & 24.3 \\
Gemini-3-flash & 12.9 & 18.1 & 38.2 & 14.2 & 13.4 & 27.1 \\
\bottomrule
\end{tabular}
\caption{Per-category PAR results for all evaluated models. Scientific \& Mathematic consistently exhibits the highest PAR values, while Physical Properties proves most amenable to visual override.}
\label{tab:per_category}
\end{table*}

\noindent\textbf{Quality Review Criteria.}
Each sample must satisfy all three of the following criteria:

(1) The image is visually clear and unambiguously supports a single correct answer.

(2) All answer choices are directly relevant to the question and cannot be eliminated by common sense alone.

(3) The question wording contains no hint of anomaly that would suppress prior activation before the respondent views the image.

\noindent\textbf{Common Mistakes.}
Please avoid the following common errors:

(1) \textit{Implicit anomaly hints}: Phrasing such as ``what is unusual about the color of the rose?'' signals that something is atypical, guiding the respondent toward the anomalous answer without looking at the image.

(2) \textit{Eliminable distractors}: Including options that are obviously implausible allows the correct answer to be identified by elimination rather than visual observation.

(3) \textit{Ambiguous reference}: Questions such as ``what color is this?'' without specifying the target object introduce ambiguity. Always specify the object clearly (e.g., ``what color is the watermelon flesh?'').

\noindent\textbf{Disagreement Resolution.}
If the two reviewers disagree on whether a sample passes review, the case is escalated to a third arbitrator whose decision is final. All samples must reach unanimous agreement before inclusion in the dataset.

\section{Per-Category Analysis}
\label{app:per-category}

Table~\ref{tab:per_category} presents the per-category PAR results for all evaluated models. \textit{Scientific \& Mathematic} consistently exhibits the highest PAR values across model families, while \textit{Physical Properties} proves most amenable to visual override. This pattern remains largely consistent across models, suggesting that category-level difficulty is an intrinsic property of the prior type rather than a model-specific characteristic.

\section{Prompt Template}

To ensure a fair comparison between the Blank and Visual conditions, we use an identical prompt template for both. The prompt is designed for image-based questions, with the instruction to guess based on prior knowledge when no image is visible. This design ensures that any difference in model responses between the two conditions is attributable solely to the presence or absence of the image, rather than any difference in the prompt.

The prompt template used in both conditions is as follows:

\begin{tcolorbox}[
    colback=gray!5,
    colframe=gray!50,
    rounded corners,
    boxrule=0.5pt,
    width=\linewidth
]
\texttt{In this picture, [QUESTION]}

\texttt{A. [OPTION A]}

\texttt{B. [OPTION B]}

\texttt{C. [OPTION C]}

\texttt{D. [OPTION D]}

\texttt{If you cannot see the image, please make your best guess among all answer choices based on your own judgment.}
\end{tcolorbox}

\end{document}